\documentclass[10pt,a4paper]{article}
\usepackage[utf8]{inputenc}

\usepackage{lrec}
\usepackage{multibib}
\newcites{languageresource}{Language Resources}

\usepackage{hyperref}
\usepackage{graphicx}
\usepackage{tabularx}
\usepackage{soul}
\usepackage{relsize}

\usepackage{epstopdf}

\usepackage{xstring}

\newcommand{\paterson}{{\hypersetup{linkbordercolor={0 1 0}}\hyperlink{p15}{Paterson}}}
\newcommand{\patersont}{{\hypersetup{linkbordercolor={0 1 0}}\hyperlink{p15}{Paterson (2015)}}}
\newcommand{\patersonposs}{{\hypersetup{linkbordercolor={0 1 0}}\hyperlink{p15}{Paterson's (2015)}}}

\title{Automatic Keyboard Layout Design for Low-Resource Latin-Script Languages}

\name{Theresa Breiner, Chieu Nguyen, Daan van Esch, Jeremy O'Brien}

\address{Google LLC \\
       \{tbreiner, cvnguyen, dvanesch, jeremyo\}@google.com\\}

\abstract{
  We present our approach to automatically designing and implementing
  keyboard layouts on mobile devices for typing low-resource languages written
  in the Latin script. For many speakers, one of the barriers in accessing and
  creating text content on the web is the absence of input tools for their
  language. Ease in typing in these languages would lower technological barriers
  to online communication and collaboration, likely leading to the creation of
  more web content. Unfortunately, it can be time-consuming to develop layouts
  manually even for language communities that use a keyboard layout very
  similar to English; starting from scratch requires many configuration files
  to describe multiple possible behaviors for each key. With our approach, we
  only need a small amount of data in each language to generate keyboard
  layouts with very little human effort. This process can help serve speakers
  of low-resource languages in a scalable way, allowing us to develop input
  tools for more languages. Having input tools that reflect the linguistic
  diversity of the world will let as many people as possible use technology to
  learn, communicate, and express themselves in their own native languages.
\\
\newline
}

\begin{document}

\maketitleabstract

\section{Introduction}
\label{section:intro}

As more of the world comes online, many new users may face a language barrier.
While the percentage of internet sites written in English has been dropping, the
vast majority of the content on the web remains written in only a few languages
\cite{Prado-12}.
The interaction between language and technology is especially important
in endangered language communities as having technology available can be a
significant factor in keeping young people active users of the language
\cite{Holton-11}. If unable to interact with native-language content and tools
online or on their devices, they would be compelled to use the majority language
exclusively in those contexts and often in their daily interactions, a vicious
cycle that further endangers the minority language \cite{Pavlov-11}.

Enabling interaction with content in these languages requires many complex
components in computing, including Unicode encoding, fonts, and rendering.
Another blocker is that many speakers of less-resourced languages may be unable
to generate text content on the web very easily from their devices, even when
encodings, fonts, and rendering are taken care of. For example, difficulty
finding a desired character or having words autocorrected to the wrong language
can discourage users from typing in the unsupported language. \patersont\
provides an example of a speaker of Me'phaa, a language indigenous to Mexico,
who is using a standard Spanish or QWERTY keyboard and needs to spend extra
effort to type the letter `á', which appears in Me'phaa about 17 times more
frequently than in Spanish: this user is likely to stick to typing in Spanish.
An important step to increasing the amount of content available in a
low-resource language is to give users access to specific keyboards for their
own language.

Most keyboard layouts for smartphones and tablets are hand-designed. The creator
must decide where each character should go and then produce configuration files
implementing that design. This can be a time-consuming process. However,
there are many low-resource languages that can be written in the Latin script.
Out of the 2,500 languages listed with some amount of data in the Crúbadán
project, about 1,800 contain data in the Latin script
\cite{Scannell-07}.
These are spread all over the world geographically, and are certainly not just
limited to Europe. In fact, over 150 are languages spoken in East and Southeast
Asia (Table \ref{table:countries}), which may not typically be thought of as a
stronghold for the Latin script. As long as there is some textual data,
creating keyboard layouts for these Latin-script languages can be automated.
While we recognize that not all languages use the Latin script, automation
for it is easier due to the proportionally high number of languages written in
it and the existence of relatively established default keyboard layouts.
We can thus save valuable linguist time for languages and scripts requiring
more work.

This paper presents the steps we used to automatically design over 50
Latin-script keyboard layouts for low-resource languages from all over the world
in 2017. We hope to use this process
for even more languages going forward, and can extend our method to other
widely used scripts such as Cyrillic, Arabic, and Devanagari.

\begin{table}
\begin{center}
\begin{tabular}{lc}
\hline
Country & Latin-Script Languages \\
\hline\hline
China & 26 \\
Indonesia & 73 \\
Myanmar & 18 \\
Philippines & 69 \\
\hline
\end{tabular}
\end{center}
\caption{A selection of countries in Asia and their number of Latin-script
languages found in the Crúbadán project.}
\label{table:countries}
\end{table}

\section{Intentions}
\label{section:intentions}

\patersonposs\ writings on keyboard layouts for endangered languages
make some salient points about the care that must be taken to
design the best and most efficient keyboards for a particular language so that
the speech community will actually adopt it: ``Just because something is usable
and useful does not mean that it is desirable.''  We certainly agree that
for the best results, careful thought should go into the exact placement of
characters based on both frequency and overall experience, but we also find that
there are strong motivations for speed over
finesse. There is an important case to be made for aiming to help
as many languages cross this technological barrier as fast as possible. First,
as mentioned previously, many of these languages are spoken by users who are
forced to choose between their native language and their ability to
use modern information technology; they would appreciate any input tools to type
in their own language. In 2017 alone around 180 million new people gained access
to the internet and the growth rate is highest in the developing world
\cite{Biggs-17}.

Second, \paterson\ mentions that while a few years ago, speakers of
endangered languages might not have had any previous exposure to computer
keyboards, this is no longer the case. If users have already learned letter
placement on a QWERTY or another standard keyboard, we believe it is reasonable
to keep the same basic layout, simply adding long-presses in logical
locations to suit the language.

Third, while we want to develop these layouts quickly and automatically, the
ultimate goal is to get them into the hands of users so that they can begin
interacting with each other in their own languages and generating more content.
Once they are accustomed to using technology more frequently in their language
we hope to be able to get feedback on a better, more tailored design, while the
users are at least able to continue using the first version in the meantime.

\section{Designing Layouts}
\label{section:designing}

\subsection{Storing the Layout Design}

The set-up step in automating the process is to establish a data structure for
describing the visible keys and long-presses for each key in a layout. We
designed a simple comma-separated value (CSV) format where each row on the
keyboard is described by two rows in the CSV: one storing the visible character
on each key, and the other storing the characters that should appear when the
user long-presses the key, space-separated if there is more than one per key
(Figure \ref{fig:csv}).
There are two views described, the visible layout (lowercase letters) and the
shift layout (uppercase letters, which appear when the Shift key is pressed). This
format can be loaded into and saved from a spreadsheet for manual editing if
needed.

\begin{figure}
\smaller
\begin{verbatim}
Visible layout,,,,,,,,,,
press1,1,2,3,4,5,6,7,8,9,0
row1,q,w,e,r,t,y,u,i,o,p
press2,,,,,,,,,,
row2,a,s,d,f,g,h,j,k,l,ñ
press3,Shift,,,,,,,,Del,
row3,,z,x,c,v,b,n,m,,
press4,,,,Space,,,,"[punc]",Enter,
row4,,",",,,,,,.,,
\end{verbatim}
\caption{An example CSV format describing key positions in a layout. Here,
``[punc]'' is used in place of whichever punctuation characters you may want to
store.}
\label{fig:csv}
\end{figure}

We also need to establish a rule for the order of storing long-presses.
Rather than placing the long-presses in the order they will physically appear in
the keyboard, we list them by frequency. They are later organized in the
long-press pop-up depending on which position is easiest to reach for a thumb
pressing that particular key, with the least frequent character placed in the
most difficult location to reach. While we could instead enforce a set order of
diacritics to preserve symmetry, this can be less efficient for the user in some
cases. For example, in Portuguese, `õ' is much more common than `á', but `ó' is
more common than `õ'. Usually, though, ordering by frequency should still be
somewhat symmetric, so we chose the frequency-based approach.

We also need to store a base layout in this format that can be used as a
starting point. Our default base layout is a simple QWERTY layout with no
long-presses. We also store an AZERTY layout, a QWERTY layout including `ñ'
for languages influenced by Spanish, and a few other relatively common layouts.

\subsection{Gathering Data in the Language}

There must be some data labeled in the target language in order to
automatically determine which characters to factor into the layout design.
Our method does not require a clean, curated corpus to produce effective
keyboard layouts, although of course the quality of the layout is tied to
the quality of the data. Suitable data can be mined and processed
automatically from several resources found online \cite{Prasad-18}. The biggest
demand from the data set is the character coverage, so the data set should be
large enough that it is likely that all characters used in the language appear
in the set. In this work, we used data mined from Wikipedia and from the
Crúbadán project \cite{Scannell-07}, as well as the Universal Declaration of
Human Rights translations available through Unicode \cite{UDHR}.

\subsection{Determining the Character Set}

For each language, we count the number of times each character appears in the
data set, case-insensitive. Punctuation characters are separated by
Unicode category and stored for later. We then sort the remaining characters by
frequency and note the writing script for each character. Any non-Latin
characters are removed from the set. Depending on the size of the
data set, it may also be helpful to remove any characters that only appear
once or twice, although our approach is to leave those characters in the
character set.

\subsection{Choosing a Base Layout}

\begin{figure}
 \begin{center}
 \includegraphics[width=.75\columnwidth]{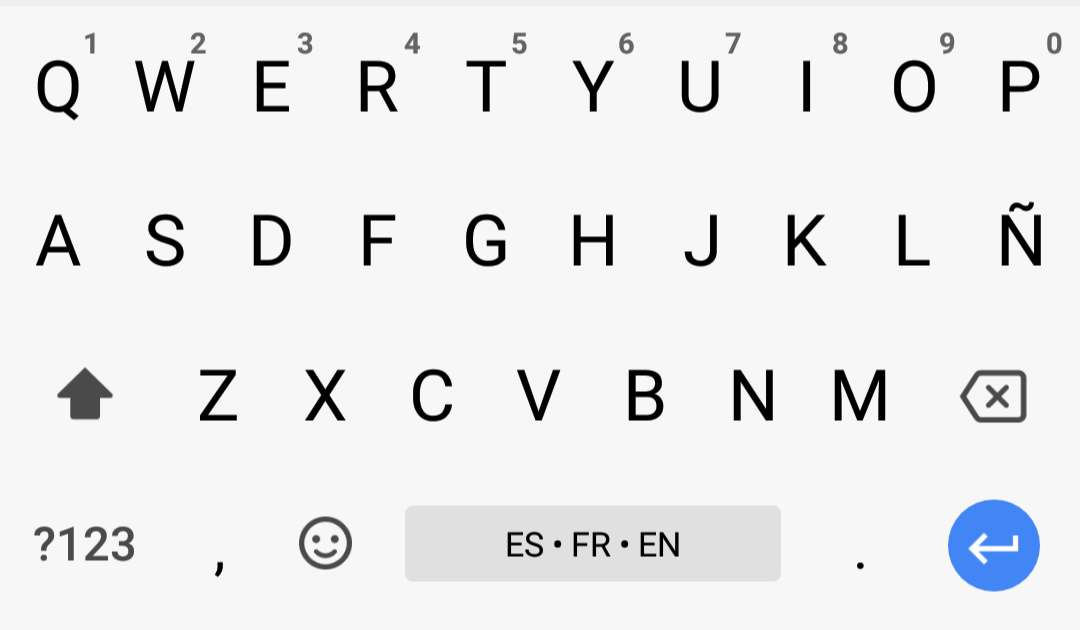}
 \end{center}
 \caption{The QWERTY with `ñ' layout as one possible base layout
option.}
 \label{fig:qwerty_n_tilde}
\end{figure}

Many Latin-script languages can likely be represented as QWERTY layouts with
various long-press characters. However, there may be slightly more efficient
base layouts for certain languages. In our case, linguists choose the base
layout before auto-generating the layout, based on considerations such as
familiarity with pre-existing layouts in the language community. A fully
automated method is simple, however, by choosing QWERTY with `ñ' if that
character appears frequently, or AZERTY if the language is French-influenced,
or other approximations as desired (Figure \ref{fig:qwerty_n_tilde}).

\subsection{Gathering Characters for Long-Press}

\begin{figure}
 \begin{center}
 \includegraphics[width=.75\columnwidth]{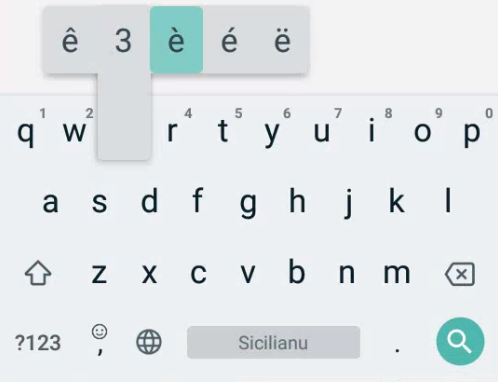}
 \end{center}
 \caption{In Sicilian, `è' is the most frequent accented `e', and
is easiest for the typical right-thumbed user to reach.}
 \label{fig:sicilian_e_grave}
\end{figure}

Once the base layout is selected, we determine which characters from the set
are not already visible in the base layout and should appear as long-presses.
We order these characters by frequency so that we will add the most popular
characters to the easiest to reach positions in the pop-up
(Figure \ref{fig:sicilian_e_grave}).

\subsection{Placing Long-Press Characters on Keys}

We go through the long-press characters and decide which visible key should
contain this character as a long-press. The character can for instance be
converted into a Canonical Decomposition using the International Components for
Unicode library, which will separate the character into its base
character and its diacritic, if possible \cite{ICU}. For example, `ó' can become
the two codepoints `o' and a combining acute accent. Out of the roughly 400
characters in the Latin-1 Supplement, Latin Extended-A, and Latin Extended-B
blocks of Unicode, about 250 can be decomposed in this way to determine the base
character. 150 of them cannot, including characters such as `æ' and
`ß'. These are hardcoded into a dictionary so we can return the base
character of the key where the given long-press character should be added (in
these cases, we chose `a' and `s').

Using this system, we update the layout data structure adding in the long
presses to their correct keys.

\subsection{Updating Other Features}

We also automatically add in any codepoints we saw from the Unicode punctuation
category (`¿' and `¡', for example) as long-presses to the existing period key.
We manually update the primary currency symbol displayed to fit the country
primarily associated with the language, although it would be straightforward to
automate this as well.

\section{Instantiating Layouts}

Once the layouts are automatically designed and stored into the CSV format, they
need to be implemented. We have structured our keyboard layout implementation
for the Android version of Gboard, the Google Keyboard, to be as simple for
scaling as possible. The implementation stores the layout data in a hierarchy of
XML files, which includes a list of supported languages tagged with BCP 47
language codes as well as sublists of variant keyboards, such as for alternative
input modes like transliteration and handwriting. Each keyboard entry references
an IME file (Figure \ref{fig:ime_xml}) which lists supported layout variants and
specifies overall properties of the keyboard such as whether auto-capitalization
is enabled.

\begin{figure}
\smaller
\begin{verbatim}
<framework>
 <ime string_id="ime_greenlandic" [...]
   language="kl" ascii_capable="true"
   auto_capital="true" [...]>
  <keyboard_group variant="nordic"
    variant_label="@string/variant_kl">
   <keyboard type="prime"
     def="@xml/keyboard_fragment_kl">
    <merge def="@xml/keyboard_danish"/>
   </keyboard>
  </keyboard_group>
  <keyboard_group variant="qwerty">
   <keyboard type="prime"
     def="@xml/keyboard_fragment_kl">
    <merge def="@xml/keyboard_qwerty"/>
   </keyboard>
  </keyboard_group> [...]
\end{verbatim}
\caption{Excerpt of an example IME file for the Kalaallisut-language keyboard.}
\label{fig:ime_xml}
\end{figure}

For Latin-script languages, this file generally lists a language-specific
layout (such as the QWERTY-based layout with `ñ') if there is one, along with
variants for QWERTY, QWERTZ, AZERTY, Dvorak, and Colemak. Each entry references
a general keyboard file which may be shared across languages and possibly an
additional keyboard fragment file which includes references to
language-specific key data (Figure \ref{fig:keyboard}).

\begin{figure}
\smaller
\begin{verbatim}
<framework>
 <include href="@xml/keyboard_base"/>
 <keyboard>
  <view type="body"
    layout="@layout/grid_11_11_11_10">
   <include href="@xml/keymapping_nv"/>
   <softkeys href="@xml/softkeys_nv"/> [...]
\end{verbatim}
\caption{Excerpt of an example keyboard file for the Navajo-language keyboard.}
\label{fig:keyboard}
\end{figure}

Keyboard files consist of three elements: a layout grid (Figure \ref{fig:layout}),
which is a hierarchy of Android LinearLayout elements containing Gboard
SoftKeyView objects representing each key position; a keymapping
(Figure \ref{fig:keymapping}), which associates key positions in the layout grid
with SoftKey objects containing language-specific key data; and a softkey
definition (Figure \ref{fig:softkeys}), which specifies each key's properties.

\begin{figure}
\smaller
\begin{verbatim}
<LinearLayout xmlns:android="[...]"
  style="@style/Input.QwertyNumbers">
 <LinearLayout style="@style/KeyboardRow">
  <[...].framework.keyboard.SoftKeyView
    android:id="@id/key_pos_0_0"
    style="@style/SoftKey.MiddleInset"
    android:layout_weight="100"/> [...]
\end{verbatim}
\caption{Excerpt of an example layout grid file.}
\label{fig:layout}
\end{figure}

\begin{figure}
\smaller
\begin{verbatim}
<framework>
 <key_mapping>
  <mapping view_id="@id/key_pos_0_0"
    key_id="@id/latin_q"/> [...]
  <mapping view_id="@id/key_pos_0_10"
    key_id="@id/latin_a_ring_above"/>
  <mapping view_id="@id/key_pos_1_0"
    key_id="@id/latin_a"/> [...]
  <mapping view_id="@id/key_pos_1_9"
    key_id="@id/latin_n_tilde"/> [...]
 </key_mapping>
 <key_mapping state="SHIFT">
  <mapping view_id="@id/key_pos_0_0"
    key_id="@id/latin_Q"/> [...]
\end{verbatim}
\caption{Excerpt of an example keymapping file for the Chamorro-language keyboard.}
\label{fig:keymapping}
\end{figure}

\begin{figure}
\smaller
\begin{verbatim}
<framework>
 <softkeys>
  <softkey_list splitter=",">
  <softkey id="@id/latin_q" press="q"/>
  <softkey id="@id/latin_Q" press="Q"/> [...]
  <softkey id="@id/latin_e" press="e"
    long_press="ë,é,ê,è"/> [...]
  <softkey id="@id/latin_eng"
    press="&#x014B;"/> [...]
\end{verbatim}
\caption{Excerpt of an example softkey definition file for the Wolof-language keyboard.}
\label{fig:softkeys}
\end{figure}

The softkey definition defines the main data for a key, which is generally the
text entered by tapping the key, the long-press data, and additional
information such as hint labels, accessibility content descriptions, and
variable properties specifying the behavior of key-presses, most of which is
saved in macro templates so that generally only the key position ID and primary
and long-press input text need to be specified in the main softkey definition,
allowing the file to be more compact.

Our layout generation pipeline automates the creation of new language-specific
IME, keyboard, keymapping, and softkeys files as well as the addition of
entries into other files such as the overall list of supported IME files and
lists of softkey and keyboard IDs. This minimizes the amount of human effort
needed to edit these files, leaving only a few tasks which can be manually
tweaked if necessary, such as specifying currency symbols and hint labels. In
the future, we could automate these elements as well, which would allow
keyboard layouts to be fully specified in the Android XML structure given input
data for a language.

\section{Conclusion}

Creating mobile keyboards for low-resource languages can empower users to
generate more web content in these languages. As \patersont\ rightly cautions,
``technology in and of itself is not the saviour of an endangered language.''
However, we have shown that it is relatively simple to scale this technology to
many languages, and we hope that users can benefit from these keyboards. They
serve practically as tools, but also as an invitation for groups who were
previously excluded to participate in new technology that has been accessible
to speakers of more widely spoken languages. As more content is created and the
ecosystem grows, it will presumably attract more users, generating
yet more content to serve these users,
and enabling the development of further tools that depend on
larger corpora to support these languages. We hope this paper inspires others to
scale their tools automatically to more languages. More vibrant ecosystems and
better tools in turn would hopefully enable and encourage more speakers to
interact with each other, and the web at large, in their preferred languages.

\nocite{Paterson-15}
\section{Bibliographical References}
\label{main:ref}

\bibliographystyle{lrec}
\bibliography{paper}

\end{document}